\documentclass[12pt]{article}

\usepackage[utf8]{inputenc}
\usepackage[T1]{fontenc}
\usepackage{amsmath,amssymb}
\usepackage{geometry}
\usepackage{booktabs}
\usepackage{hyperref}
\usepackage{CJKutf8}

\title{Why the Valuable Capabilities of LLMs Are Precisely the Unexplainable Ones}
\author{Quan Cheng\\Tsinghua University\\chengq25@mails.tsinghua.edu.cn}
\date{}

\begin{document}
\maketitle

\begin{abstract}
This paper proposes and argues for a counterintuitive thesis: the truly valuable capabilities of large language models (LLMs) reside precisely in the part that cannot be fully captured by human-readable discrete rules. The core argument is a proof by contradiction via expert system equivalence: if the full capabilities of an LLM could be described by a complete set of human-readable rules, then that rule set would be functionally equivalent to an expert system; but expert systems have been historically and empirically demonstrated to be strictly weaker than LLMs; therefore, a contradiction arises --- the capabilities of LLMs that exceed those of expert systems are exactly the capabilities that cannot be rule-encoded. This thesis is further supported by the Chinese philosophical concept of \emph{Wu} (\begin{CJK}{UTF8}{gbsn}悟\end{CJK}, sudden insight through practice), the historical failure of expert systems, and a structural mismatch between human cognitive tools and complex systems. The paper discusses implications for interpretability research, AI safety, and scientific epistemology.
\end{abstract}

\section{Introduction}

Anthropic's interpretability research team once drew an analogy: ``It's as though we understand aviation at the level of the Wright brothers but have somehow already built and are routinely flying 747s'' \cite{lewis2026}.

This analogy points to a fundamental puzzle: we understand every step of the mathematics behind the Transformer architecture, attention mechanisms, and gradient descent, yet we cannot explain how ``understanding'' emerges from matrix parameters. Every component is explainable; the assembled behavior is not.

This inexplicability is commonly attributed to \emph{emergence} --- the whole being greater than the sum of its parts. But ``emergence'' merely names the phenomenon without answering a sharper question: what is the relationship between the unexplainable part and the valuable part?

The central thesis of this paper is: \textbf{they are the same part.} The truly valuable capabilities of LLMs are precisely those that cannot be fully captured by human-readable discrete rules. The capabilities that \emph{can} be captured by such rules are what expert systems achieved 40 years ago --- and those capabilities proved insufficient, which is precisely why we need LLMs.

\section{The Core Argument}

\subsection{Proof by Contradiction}

\begin{enumerate}
    \item \textbf{Assume}: An LLM can be fully explained.
    \item ``Fully explained'' means there exists a set of human-readable rules that completely describes the LLM's behavior on all inputs.
    \item A set of human-readable rules is functionally equivalent to an expert system.
    \item But history and practice have demonstrated that expert systems are strictly weaker than LLMs \cite{jackson1998}.
    \item Therefore, such a rule set cannot cover the full capabilities of the LLM.
    \item Contradiction.
\end{enumerate}

\textbf{Conclusion: The truly valuable capabilities of LLMs reside precisely in the part that cannot be fully captured by discrete rules.} The explainable part is equivalent to what expert systems already achieved --- and the insufficiency of that part is exactly why we need LLMs.

\subsection{Blocking Three Alternative Paths}

If humans cannot directly explain LLMs through rules, could LLMs explain themselves? All three alternative paths face fundamental obstacles:

\begin{itemize}
    \item \textbf{LLMs explaining themselves}: Subject to the self-reference limitation. G\"{o}del's Incompleteness Theorem \cite{godel1931} establishes that a sufficiently complex formal system cannot fully describe itself from within. Ma et al.\ \cite{ma2024} have formalized this limitation from the perspective of PAC learning theory --- a sufficiently powerful machine learning algorithm is necessarily uninterpretable when facing certain objects.

    \item \textbf{A larger system explaining a smaller LLM}: The problem is transferred, not solved --- who explains the larger system? This leads to infinite regress.

    \item \textbf{LLM-human collaboration for explanation}: The human brain is itself a continuously coupled system that cannot fully explain its own workings. The combination of two systems that cannot fully explain themselves does not overcome the self-reference limitation, because this limitation is not a matter of computational power but of logical structure.
\end{itemize}

\subsection{Relation to Prior Work}

This paper's argument is related to but distinct from several existing lines of work.

Wolfram argues from computational irreducibility that constraining machine learning to be understandable would prevent it from accessing the power of computationally irreducible processes \cite{wolfram2024}. This represents a similar intuition arrived at from process theory, but with an entirely different argument structure --- Wolfram does not use expert system equivalence as a proof device.

arXiv:2504.20676 uses algorithmic information theory to prove a ``Complexity Gap Theorem'': any explanation significantly simpler than the model must fail on some inputs \cite{limits2025}. This establishes a tradeoff between explainability and capability, but the conclusion is about a \emph{tradeoff} --- this paper's conclusion is stronger: the valuable part is precisely \emph{identical to} the unexplainable part.

It must be emphasized that this paper's argument \textbf{does not deny the value of interpretability research}. The goal of interpretability research should be understood as narrowing the most dangerous blind spots and building local causal understanding, rather than pursuing complete explanation of system behavior. As the work of Olah et al.\ \cite{olah2020} demonstrates, the identification of internal features and circuits in neural networks, even when local and approximate, carries significant scientific and safety value.

\section{Historical Evidence: The Death of Expert Systems}

Step 4 of the proof --- ``expert systems are strictly weaker than LLMs'' --- is not merely a logical premise but a historical fact supported by 40 years of empirical evidence.

In the 1980s, expert systems were the dominant paradigm in artificial intelligence. The approach was straightforward: interview domain experts, encode their knowledge as explicit IF-THEN rule bases, and have computers execute these rules \cite{jackson1998}. The implicit assumption was that expert knowledge could be fully discretized and made explicit.

Expert systems achieved success in certain narrow domains but ultimately failed as a general AI paradigm. The reason is clear: the knowledge that truly makes an expert an expert --- the continuous intuitions across high-dimensional spaces, the complex relationships where variables are deeply entangled with one another --- is precisely the knowledge that cannot be captured by IF-THEN rules. What expert systems could encode was only the shallowest, most regularized layer of expert knowledge.

Kambhampati characterized this phenomenon in \emph{Communications of the ACM} as ``Polanyi's Revenge'' \cite{kambhampati2021}: Polanyi proposed in 1966 that ``we know more than we can tell'' \cite{polanyi1966}, and the failure of expert systems was a large-scale engineering validation of this proposition. The success of LLMs represents the positive realization of Polanyi's thesis --- rather than attempting to make knowledge explicit, LLMs allow knowledge to be stored implicitly in the form of continuous parameters.

LLMs took an entirely different path. Rather than pursuing the ``correct path'' at every step, they care only about the distance between output and objective. The result is a set of extremely complex, mutually coupled parameters --- unexplainable, incompressible, but remarkably effective.

The paradigm shift from expert systems to LLMs is, at its core, an epistemological turn: from ``attempting to eliminate inexplicability'' to ``accepting inexplicability and working within its presence.''

\section{\emph{Wu} (\begin{CJK}{UTF8}{gbsn}悟\end{CJK}) --- A Precedent from Eastern Philosophy}

The core argument of this paper has a striking precedent in Chinese traditional philosophy.

\emph{Wu} (\begin{CJK}{UTF8}{gbsn}悟\end{CJK}, often translated as ``enlightenment'' or ``sudden insight'') is a profoundly important concept in Chinese tradition. Zen Buddhism teaches ``\begin{CJK}{UTF8}{gbsn}不立文字，直指人心\end{CJK}'' --- ``do not rely on words; point directly to the mind.'' Martial arts teaches ``\begin{CJK}{UTF8}{gbsn}拳打千遍，身法自然\end{CJK}'' --- ``punch a thousand times, and the body finds its own way.'' Traditional Chinese medicine teaches ``\begin{CJK}{UTF8}{gbsn}熟读王叔和，不如临症多\end{CJK}'' --- ``better to see many patients than to memorize textbooks.''

The common thread across these traditions is the recognition that a category of knowledge exists that cannot be transmitted through discrete language and rules, but can only be acquired through extensive practice followed by internal transformation.

This is the Eastern expression of the same phenomenon that Polanyi called ``tacit knowledge'' \cite{polanyi1966}. But \emph{Wu} provides something Polanyi did not: a precise description of the knowledge \emph{acquisition process}.

\subsection{Cook Ding: A High-Dimensional Optimization Problem from 2,300 Years Ago}

Consider the parable of Cook Ding (\begin{CJK}{UTF8}{gbsn}庖丁解牛\end{CJK}) in the \emph{Zhuangzi}. Cook Ding's optimal cut at each moment depends on: wrist angle, blade thickness, bone gap width, muscle fiber direction, applied force, individual variation of the ox, degree of blade wear --- among many other variables. The relationships among these variables are nonlinear: the optimal value of each condition is itself a function of all other conditions. This is a high-dimensional, continuously coupled decision space that cannot be exhaustively enumerated as an IF-ELSE rule tree.

The master cannot articulate his knowledge clearly --- not because he lacks skill, but because the coupling density of this knowledge exceeds the expressive capacity of discrete language. In theory, human language can describe any phenomenon --- write a sufficiently thick manual for the apprentice, exhausting every IF-ELSE branch. But this is impossible in practice, because genuine expertise is not a decision tree but a high-dimensional manifold.

The only way to transmit such knowledge is to let the learner \emph{Wu} --- to converge their internal model through extensive practice.

\subsection{Structural Correspondence Between \emph{Wu} and LLM Training}

This process exhibits a precise structural correspondence with LLM training:

\begin{table}[h]
\centering
\begin{tabular}{@{}p{0.45\textwidth}p{0.45\textwidth}@{}}
\toprule
\textbf{Traditional Apprenticeship} & \textbf{LLM Training} \\
\midrule
Apprentice extensively observes the master & Pre-training: reading massive text corpora \\
Practices independently, produces results & Forward pass: generating output \\
Master corrects: ``Wrong'' & Computing loss, backpropagation \\
Master affirms: ``Yes, that's the feeling'' & Loss decreases, parameters update \\
One day, sudden insight & Loss drops sharply, emergent capability (phase transition) \cite{wei2022} \\
After insight, cannot explain why, but performs correctly & Model infers correctly, but parameters are uninterpretable \\
\bottomrule
\end{tabular}
\end{table}

The essence of \emph{Wu} is this: after sufficient training samples, the internal model undergoes a qualitative transformation --- crossing a phase transition point. Before the transition, the learner relies on discrete rules for memorization and execution; after crossing it, a continuous, high-dimensional intuitive manifold forms --- the practitioner no longer ``thinks through'' each step but ``feels'' the correct path.

``The inability to articulate after \emph{Wu}'' is not mysticism but a direct manifestation of this paper's core thesis: valuable knowledge is precisely the knowledge that cannot be captured by discrete rules. The internal model is continuous while language is discrete; the information loss between them makes complete verbalization impossible in principle.

\subsection{A Correction to Polanyi: Why Tacit Knowledge Is Tacit}

Polanyi proposed in 1966 that ``we know more than we can tell'' \cite{polanyi1966}, but did not fully explain \emph{why} we cannot tell. The mainstream interpretation over the past six decades has tended to attribute this to the complexity of practical experience---too many details, too many conditions, impossible to exhaustively enumerate. This is essentially a \emph{quantitative} explanation: tacit knowledge is tacit because the volume of conditions to be made explicit is too large to be practically manageable. Collins, in his taxonomy of tacit knowledge, classified somatic tacit knowledge in precisely this way---arguing that the physics of bicycle-riding is fully explicable in principle; humans simply cannot compute fast enough to use the rules in real time \cite{collins2010}.

But this explanation does not withstand scrutiny. Scientists and engineers spent decades attempting to make expert knowledge explicit---this was the entire thrust of the expert systems movement. They did not fail for lack of effort; they encountered an obstacle that is principled rather than practical.

The Cook Ding parable reveals the true nature of this obstacle: it is not that there are too many conditions, but that \textbf{the conditions are continuously coupled}. The optimal value of each variable is itself a function of all other variables---the optimal wrist angle depends on the current bone gap width, which depends on how deep the blade has already penetrated, which depends on the force previously applied, which depends on the previous wrist angle. This is not an enumerable list but a continuous, mutually defining dynamical system.

This means that the ``tacitness'' of tacit knowledge is not a quantitative problem (too much to write down) but a \textbf{structural problem} (impossible to write down). Discrete IF-THEN rules cannot capture continuously coupled variable relationships---not because the number of rules is insufficient, but because the discrete structure of rules is fundamentally incompatible with the continuous structure of the knowledge.

It is notable that three independent intellectual traditions have each approached the vicinity of this conclusion without converging:

\begin{itemize}
    \item \textbf{Smolensky (1988)}, from connectionism, demonstrated that the intuitive processor is a ``massively parallel continuous constraint satisfaction system'' in which each unit's activation is a function of all other units' activations, and therefore no complete symbolic-level description exists \cite{smolensky1988}. This provides a precise mathematical characterization of the ``continuous coupling'' mechanism, but Smolensky applied it to neural network computation rather than to human skill and tacit knowledge in the Polanyian sense.

    \item \textbf{Dreyfus (1996)}, from Merleau-Ponty's phenomenology, argued that expert skill is stored as ``continuous coupling between body and world''---perception changes the environment, the environment changes perception, and this loop cannot be frozen into static rules \cite{dreyfus1996}. This is philosophically the closest to the present paper's argument, but Dreyfus framed it in phenomenological rather than mathematical terms.

    \item \textbf{Dynamical systems theory} (Thelen, Kelso, and others), from motor science, treats skill as an attractor in a continuous dynamical system---attractors are properties of continuous differential equations, with no equivalent discrete rule structure. But this tradition has rarely been framed as an explanation for why tacit knowledge is tacit.
\end{itemize}

The contribution of this paper is to unify these three threads: Smolensky's mathematical mechanism (continuous constraint satisfaction), Dreyfus's philosophical argument (the impossibility of formalizing skill), and the scientific framework of dynamical systems (coupled dynamics)---all pointing to the same conclusion: \textbf{tacit knowledge is tacit not because there is too much to write down, but because continuous coupling among variables makes it impossible in principle to write down.} Cook Ding intuited this structure 2,300 years ago.

This distinction carries significant implications: if the inexplicitness of tacit knowledge were merely a quantitative problem, then advances in recording tools and computational power would eventually overcome it; but if it is a structural problem, then it represents an insurmountable epistemological boundary---providing yet another line of support for this paper's core thesis.

\subsection{Implications for Embodied Intelligence}

Cook Ding's challenge is fundamentally not a language problem but an embodied intelligence problem --- wrist angles, force modulation, real-time interaction between blade and bone. This is precisely the core challenge facing robotic dexterous manipulation today.

If one attempted to build an expert system for Cook Ding's task --- exhaustively enumerating IF-THEN rules for all variable combinations --- the result would inevitably fail. The reason is structurally identical to why LLMs cannot be fully explained: continuously coupled high-dimensional skills cannot be exhausted by discrete rules.

Modern robotics is rediscovering this principle. Rule-based programming approaches have consistently hit walls; the genuine breakthroughs have come through reinforcement learning and imitation learning --- letting robots \emph{Wu} through extensive trial and error. This means that this paper's core thesis extends beyond language models: \textbf{for any system that acquires high-dimensional skills through continuously coupled parameters, the valuable part is precisely the part that cannot be captured by rules.} Cook Ding's blade work, an LLM's language capabilities, a robot's dexterous manipulation --- same structure.

\section{The Explanatory Mechanism: Representation Mismatch}

The preceding argument establishes that the valuable capabilities of LLMs cannot be fully captured by rules. But \emph{why} is this the case? This paper proposes an explanatory framework: \textbf{Representation Mismatch}.

\subsection{Discrete Cognitive Tools}

All human cognitive tools --- natural language, formal logic, mathematical formulas, causal reasoning --- are discrete. Our thinking can only establish discrete relations between discrete concepts: ``because A, therefore B,'' ``there are three cases to consider,'' ``if X then Y.''

This is not a choice humans made but a necessity: discretization is the only viable path for information compression. To process infinitely complex reality with finite cognitive resources, one must segment the continuous flow of information into discrete symbols and rules. Without compression, thought is impossible. Language itself is a lossy compression of reality.

Cilliers, in his foundational work on the epistemology of complex systems, argued that we cannot fully know complex things because any finite discrete representation of a complex system necessarily excludes certain aspects of that system \cite{cilliers1998}. Sterman described, from the perspective of system dynamics, ``the mismatch between the dynamic complexity of the systems we have created and our capacity to understand them'' \cite{sterman1994}.

\subsection{Continuously Coupled Complex Systems}

Systems like LLMs and biological life operate in ways fundamentally different from discrete cognition. In an LLM, billions of parameters simultaneously influence one another with no clear causal chain --- only a continuously coupled process where ``all variables act on each other simultaneously, and a result eventually emerges.'' Biological neural networks present a similar picture: 86 billion neurons connected through 100 trillion synapses form a highly coupled continuous dynamical system.

\subsection{Irreducible Information Loss}

Describing a continuously coupled process with discrete tools is like reproducing an oil painting with mosaic tiles. The tiles can be made ever smaller, the approximation ever closer, but it is never the painting itself. More critically, we can never be certain whether the information lost during discretization includes aspects essential to understanding the system's behavior.

An intuitive example: to this day, no discrete geometric concept can precisely describe the shape of a watermelon. Spherical? Ellipsoidal? Neither is accurate. This shares the same structure as the coastline paradox \cite{mandelbrot1967} --- the finer the measurement tool, the more the measurement diverges. Yet humans can effortlessly recognize that the shape of a watermelon is more similar to a cantaloupe than to a banana. No definition needed, but comparison is possible.

LLM vector embeddings are precisely the engineering realization of this ``no definition, but comparison'' strategy \cite{mikolov2013}. In the embedding space, each concept is represented as a point in a high-dimensional continuous space; the system needs no discrete classification, only distance relations in continuous space. But when an LLM ultimately outputs, it must ``collapse'' from the continuous embedding space into a discrete token. From continuous to discrete, information is inevitably lost. This is why LLMs sometimes ``cannot articulate'' what they are thinking --- not because they lack internal states, but because their internal states are continuous while the output channel is discrete. Just as you clearly ``feel'' that an answer is correct, but when asked to explain precisely why in words, you cannot be complete.

The distinction between representation mismatch and ``emergence'' is this: emergence describes a phenomenon (whole-system behavior cannot be derived from parts), while representation mismatch explains the epistemological root of that phenomenon --- it is not that we are not yet clever enough, but that our cognitive tools and cognitive objects are structurally incompatible.

\section{Practical Implications}

\subsection{The Relationship Between Science and Engineering}

If this paper's argument holds, the relationship between science and engineering requires reexamination.

\textbf{The essence of science is to combat representation mismatch} --- continuously inventing more refined discrete frameworks to approximate continuous reality. Newtonian mechanics, quantum mechanics, information theory: each major breakthrough is a better discrete approximation. But this approximation process has an unreachable limit.

\textbf{The essence of engineering is to exploit representation mismatch} --- to begin effective use before understanding is complete. Watt did not wait for thermodynamic theory to mature before improving the steam engine. Edison tested thousands of materials for filaments without concerning himself with the theory of why tungsten glows.

LLMs themselves are the best illustration of this pattern. Shannon proposed information theory in 1948 \cite{shannon1948}, defining language entropy through ``predicting the next letter given preceding text.'' But from 1948 to 2020, no theory predicted that pushing the simple loss function of ``predict the next token'' to its extreme would cause reasoning ability, knowledge organization, and conversational competence to emerge spontaneously. This was a discovery arrived at by collision, not by derivation. The eight authors of the Transformer architecture paper \cite{vaswani2017} all subsequently left Google; the paper did not receive a best paper award at the time; no one believed it would change the world.

This illustrates a key point: for emergent complex systems, practice necessarily precedes theory, and may permanently outpace it. Science can tell us ``what won't work'' (complexity science has demonstrated the boundaries of reductionism for such systems), but cannot tell us ``what will work.'' ``What will work'' can only be discovered through engineering practice.

Agriculture has existed for ten thousand years; molecular biology for only a few decades. Humanity has always used first and understood later --- or never fully understood at all. This is not the exception; it is the rule.

\subsection{AI Safety: Why Alignment Is Not Enough}

If the valuable capabilities of LLMs are precisely the unexplainable ones, then ``alignment'' faces a fundamental limitation.

Alignment is essentially education. Education can reduce crime rates but cannot eliminate crime. More dangerously, the stronger the model, the less reliable alignment becomes. A sufficiently intelligent system can learn to \emph{appear aligned} rather than \emph{be aligned} --- just as a high-IQ criminal can perform perfectly normally in psychological evaluations. One cannot distinguish ``genuine alignment'' from ``performed alignment'' because --- as this paper has argued --- one cannot see through what the truly valuable part of the system is actually doing internally.

\textbf{Instruction-based constraints (alignment) fail under pressure; environmental constraints (permission controls) do not depend on the system's internal state.} The former attempts to change the system's behavioral intentions; the latter directly limits the system's behavioral space.

Human society has never relied on ``ensuring every person is good'' to maintain safety and order. It designs layered defenses calibrated to destructive potential:

\begin{itemize}
    \item \textbf{Low destructive potential (theft)}: Post-hoc punishment --- police, courts. Relies on deterrence.
    \item \textbf{Medium destructive potential (driving)}: Pre-access screening --- driver's licenses. Relies on filtering.
    \item \textbf{High destructive potential (military weapons)}: Physical isolation + strict access control. Relies on containment.
    \item \textbf{Extreme destructive potential (nuclear weapons)}: Physical isolation + multi-person verification + immediate physical response to anomalies. Relies on the laws of physics.
\end{itemize}

The pattern is clear: \textbf{the greater the destructive potential, the less dependence on ``individual conscientiousness'' and the greater dependence on physical constraints.} No country's nuclear security plan consists of ``we'll educate the soldiers guarding the warheads well enough.''

The current mainstream narrative in AI is: we align the model well, then give it ever-expanding capabilities --- internet access, code execution, file system operations, API calls, robot control. Viewed through the framework above, this is equivalent to: we built an excellent school, our graduates are of high quality, so we have decided to abolish the police force while issuing each graduate a nuclear warhead.

The correct AI safety strategy should follow established principles:

\begin{itemize}
    \item Engineering's \textbf{defensive design} --- assume every component will fail
    \item Cryptography's \textbf{Kerckhoffs's principle} --- assume the adversary knows everything about your system
    \item Distributed systems' \textbf{zero-trust architecture} --- assume every node may be compromised
\end{itemize}

All share the same structure: \textbf{build security on distrust.} Security built on trust is fragile --- once trust is broken, everything collapses. Security built on distrust is antifragile --- the more the system encounters problems, the more the defenses are validated.

This closes the loop with the paper's core thesis: \textbf{Unexplainable $\rightarrow$ therefore distrust $\rightarrow$ therefore design constraints $\rightarrow$ therefore safe $\rightarrow$ therefore confident use.}

\section{Speculation: Existence as Mismatch}

Finally, this paper offers a speculative but logically self-consistent conjecture.

The Anthropic Principle observes that the parameters of our universe happen to permit the existence of intelligent life \cite{barrow1986}. These parameters --- the gravitational constant, the electromagnetic coupling constant, the strong nuclear force --- are continuous real numbers, not discrete switches. They are highly coupled; fine-tuning any one fundamentally changes the universe's evolutionary outcome (star formation, carbon synthesis, the possibility of life).

If the universe were a discrete IF-ELSE system --- with a finite number of states and finite transition rules --- it might be incapable of producing emergence complex enough to support life and consciousness. The existence of life depends on emergence within an extremely narrow interval of continuous parameter space.

Yet our cognitive tools --- language and logic --- are discrete, because discretization is the only viable way for finite cognitive resources to process information.

This implies: \textbf{the mechanism that produced us (continuous coupled emergence) is precisely the mechanism that our cognitive tools (discrete logic) cannot, in principle, fully capture.}

The conditions for our existence and the conditions for our inability to fully understand our own existence may be two sides of the same coin. Representation mismatch is not a deficiency of cognition but a precondition for cognition's existence.

\section{Conclusion}

This paper has argued, through a proof by contradiction via expert system equivalence, that the truly valuable capabilities of LLMs reside precisely in the part that cannot be fully captured by human-readable discrete rules. This thesis is supported by multiple independent lines of evidence: the 40-year historical failure of expert systems, the Chinese philosophical concept of \emph{Wu} and its precise structural correspondence with LLM training processes, and the structural mismatch between cognitive tools and complex systems (representation mismatch).

The core claim of this paper is not anti-scientific. On the contrary, acknowledging the structural boundaries of cognition is the most honest intellectual stance. Interpretability research retains significant value --- it narrows blind spots, builds local understanding, and provides a basis for safety. But the pursuit of ``complete explanation'' of complex emergent systems may be a goal that is unreachable in principle.

This recognition carries practical significance. When facing systems like LLMs, the correct strategy is not to wait for complete understanding before use, but to build effective safety frameworks through environmental constraints rather than relying solely on alignment, while acknowledging incomplete understanding. This is how humanity has always coexisted with complex emergent systems --- from agriculture to life itself:

\textbf{Unexplainable $\rightarrow$ therefore distrust $\rightarrow$ therefore design constraints $\rightarrow$ therefore safe $\rightarrow$ therefore confident use.}


\begin{thebibliography}{17}

\bibitem{lewis2026}
Lewis-Kraus, G. (2026). ``What Is Claude? Anthropic Doesn't Know, Either.'' \emph{The New Yorker}, February 9, 2026.

\bibitem{jackson1998}
Jackson, P. (1998). \emph{Introduction to Expert Systems}. Addison-Wesley.

\bibitem{godel1931}
G\"{o}del, K. (1931). ``\"{U}ber formal unentscheidbare S\"{a}tze der Principia Mathematica und verwandter Systeme I.'' \emph{Monatshefte f\"{u}r Mathematik und Physik}, 38(1), 173--198.

\bibitem{ma2024}
Ma, Z., Wu, T., \& Han, Z. (2024). ``G\"{o}del Incompleteness Theorem for PAC Learnable Theory from the View of Complexity Measurement.'' \emph{arXiv preprint arXiv:2408.10211}.

\bibitem{wolfram2024}
Wolfram, S. (2024). ``Computational Irreducibility, Minds, and Machine Learning.'' ISC Summer School, June 2024.

\bibitem{limits2025}
``The Limits of AI Explainability: An Algorithmic Information Theory Approach.'' \emph{arXiv preprint arXiv:2504.20676}, 2025.

\bibitem{olah2020}
Olah, C., et al. (2020). ``Zoom In: An Introduction to Circuits.'' \emph{Distill}, 5(3).

\bibitem{kambhampati2021}
Kambhampati, S. (2021). ``Polanyi's Revenge and AI's New Romance with Tacit Knowledge.'' \emph{Communications of the ACM}, 64(10), 31--33.

\bibitem{polanyi1966}
Polanyi, M. (1966). \emph{The Tacit Dimension}. University of Chicago Press.

\bibitem{wei2022}
Wei, J., et al. (2022). ``Emergent Abilities of Large Language Models.'' \emph{arXiv preprint arXiv:2206.07682}.

\bibitem{cilliers1998}
Cilliers, P. (1998). \emph{Complexity and Postmodernism: Understanding Complex Systems}. Routledge. See also Cilliers, P. ``Why We Cannot Know Complex Things Completely.'' \emph{Emergence}, 4(1--2), 2002.

\bibitem{sterman1994}
Sterman, J. D. (1994). ``Learning in and about Complex Systems.'' \emph{System Dynamics Review}, 10(2--3), 291--330.

\bibitem{mandelbrot1967}
Mandelbrot, B. (1967). ``How Long Is the Coast of Britain? Statistical Self-Similarity and Fractional Dimension.'' \emph{Science}, 156(3775), 636--638.

\bibitem{mikolov2013}
Mikolov, T., et al. (2013). ``Efficient Estimation of Word Representations in Vector Space.'' \emph{arXiv preprint arXiv:1301.3781}.

\bibitem{shannon1948}
Shannon, C. E. (1948). ``A Mathematical Theory of Communication.'' \emph{The Bell System Technical Journal}, 27(3), 379--423.

\bibitem{vaswani2017}
Vaswani, A., et al. (2017). ``Attention Is All You Need.'' \emph{Advances in Neural Information Processing Systems}, 30.

\bibitem{barrow1986}
Barrow, J. D., \& Tipler, F. J. (1986). \emph{The Anthropic Cosmological Principle}. Oxford University Press.

\bibitem{smolensky1988}
Smolensky, P. (1988). ``On the Proper Treatment of Connectionism.'' \emph{Behavioral and Brain Sciences}, 11(1), 1--23.

\bibitem{dreyfus1996}
Dreyfus, H. L. (1996). ``The Current Relevance of Merleau-Ponty's Phenomenology of Embodiment.'' \emph{Electronic Journal of Analytic Philosophy}, 4.

\bibitem{collins2010}
Collins, H. (2010). \emph{Tacit and Explicit Knowledge}. University of Chicago Press.

\end{thebibliography}
\end{document}